\documentclass[letterpaper, 10 pt, conference]{ieeeconf}  

\IEEEoverridecommandlockouts                              

\usepackage{caption}
\usepackage[skins]{tcolorbox}  
\usepackage{subcaption}
\usepackage{multicol}

\usepackage{mathtools}
\usepackage{graphics} 
\usepackage{epsfig} 
\usepackage{mathptmx} 
\usepackage{times} 
\usepackage{amsmath} 
\usepackage{amssymb}  
\usepackage{graphicx}
\usepackage{epstopdf}
\usepackage{pdfpages}
\usepackage{cite}
\usepackage{amsbsy}
\usepackage{bm}
\usepackage{url}
\usepackage{accents}

\pdfoutput=1
\usepackage{hyperref}
\hypersetup{
    colorlinks=true,
    linkcolor=black,
    citecolor=black,
    filecolor=black,
    urlcolor=black,
}

\def\horizontaldistance{\kern2pt}

\title{\LARGE \bf
SURENA\hspace{0.07cm}\rom{4}: Towards  A Cost-effective Full-size Humanoid \\ Robot for Real-world Scenarios
}

\author{Aghil Yousefi-Koma$^{1,\dagger}$, Behnam Maleki$^{1,\dagger}$, Hessam Maleki$^{1,\dagger}$, Amin Amani$^{1,\dagger}$\\
Mohammad Ali Bazrafshani$^{1,\dagger}$ Hossein Keshavarz$^{1,\dagger}$,Ala Iranmanesh $^{1,\dagger}$, Alireza Yazdanpanah$^{1,\dagger}$ \\
Hamidreza Alai$^{1,\dagger}$,\hspace{0.11cm} Sahel Salehi$^{1,\dagger}$,\hspace{0.11cm} Mahyar Ashkvari$^{1,\dagger}$,\hspace{0.11cm} Milad Mousavi$^{1,\dagger}$,\hspace{0.11cm} and Milad Shafiee Ashtiani$^{1,\dagger}$
\thanks{$\dagger$ All authors equally contributed to the paper.}
\thanks{$^{1}$Center of Advanced Systems and Technologies (CAST) School of
Mechanical Engineering, College of Engineering, University of Tehran,
Tehran, Iran.
        {\tt\small aykoma@ut.ac.ir}}%

 }

\makeatletter
\newcommand*{\rom}[1]{\expandafter\@slowromancap\romannumeral #1@}
\makeatother
\begin{document}

\maketitle
\thispagestyle{empty}
\pagestyle{empty}

\begin{abstract}
This paper describes the hardware, software framework, and experimental testing of SURENA \rom{4} humanoid robotics platform.
SURENA \rom{4} has 43 degrees of freedom (DoFs), including seven DoFs for each arm, six DoFs for each hand, and six DoFs for each leg, with a height of 170 cm and a mass of 68 kg and morphological and mass properties similar to an average adult human. SURENA \rom{4} aims to realize a cost-effective and anthropomorphic humanoid robot for real-world scenarios. In this way, we demonstrate a locomotion framework based on a novel and inexpensive predictive foot sensor that enables walking with 7cm foot position error because of accumulative error of links and connections' deflection(that has been manufactured by the tools which are available in the Universities). Thanks to this sensor, the robot can walk on unknown obstacles without any force feedback, by online adaptation of foot height and orientation. Moreover, the arm and hand of the robot have been designed to grasp the objects with different stiffness and geometries that enable the robot to do drilling, visual servoing of a moving object, and writing his name on the white-board.

\end{abstract}

\section{INTRODUCTION}

In recent years, many accomplishments have been carried out in the field of humanoid robotics. The humanoid robots such as Atlas from Boston Dynamics and Asimo from Honda are two frontiers, however, there is not technical information available about these humanoid robots which highly rely on two famous companies with massive sources of funding. The other full-size humanoid robots such as Hubo\cite{jung2018development}, HRP robots family\cite{Kaneko2011,kaneko2019humanoid}, Jaxon\cite{kojima2015development}, iCub\cite{Tsagarakis2007}, Walkman\cite{Tsagarakis2017}, Valkyrie\cite{radford2015valkyrie}, Digit\cite{digit}, TORO\cite{englsberger2014overview}, Talos\cite{stasse2017talos}   are in a same level of promising performances, so-called \textit{classic} full-size humanoid robots. 

All of these classic full-size humanoid robots have been developed by collaboration with industrial companies\cite{englsberger2014overview, Tsagarakis2017,kaneko2019humanoid, stasse2017talos}. They were made by advanced manufacturing systems to be rigid enough and have a minimum deflection for applying the rigid body dynamics algorithms for locomotion planning and control. The mechanical links of humanoid robots that are built by students or by small manufacturers are made by bending or cutting and the whole structure is fragile. By contrast, the classic humanoid robots use cast mechanical links with high rigidity and lightweight using the most advanced mechanical CAD. It was obvious that the cast links should have such properties, but the links were too expensive to be developed by university projects\cite{kajita2014introduction}.

Moreover, the conventional position-controlled humanoid robots include 6-axis Force/Torque sensors at the end-effectors of the legs and arms to control the forces and torques\cite{jung2018development,Kaneko2011,kaneko2019humanoid,jeong2020design, buschmann2009humanoid}. Additionally, recently developed torque-controlled humanoid robots include rotary torque sensors at each joints\cite{ englsberger2014overview,stasse2017talos} which are  expensive.
Therefore, the  Force/Torque sensors and the quality of manufacturing are two important characteristics that will increase the cost of humanoid robots drastically. For example, the cost of the Talos humanoid robot developed recently by Pal-Robotics company is about one million Euro.\footnote{https://spectrum.ieee.org/automaton/robotics/humanoids/talos-humanoid-now-available-from-pal-robotics}
\begin{figure}[]
\centering
\includegraphics[scale=0.33, trim ={0.0cm 2.5cm 0.0cm 2.0cm},clip]{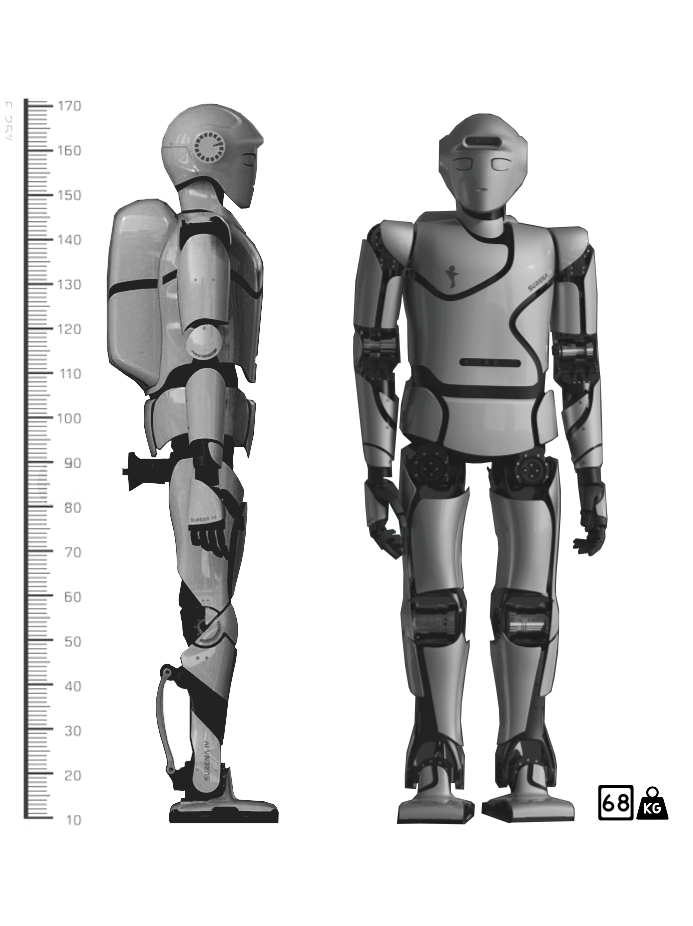}
\caption{The SURENA \rom{4} humanoid robot}
      \label{fig:surena4}
\end{figure}

Although the force-torque sensors and high-quality manufacturing could be necessary for exploiting the whole dynamic of the system in disaster scenarios and athlete-type scenarios\cite{kamioka2018simultaneous}, there are many real-world scenarios that humanoid robots do not need aggressive locomotion behaviors. As an example, consider the scenarios in which there are narrow corridors and the humanoid robot should pick and place some objects from one place in a factory to another room by climbing stairs. While the advancement of recent years is showing conventional humanoid robots will be able to do such scenarios in near future\cite{kheddar2019humanoid}, the main challenge for employing these robots is their high cost\cite{kopacek2011cost} and it is the main reason that there exist a few research institutions that are developing humanoid robots.

In this paper, we present SURENA \rom{4}, a full-size humanoid robot that has been targeted for real-world scenarios under uncertainty. SURENA does not have Force/Torque sensors and has been developed by a non-expensive manufacturing process that is available in the universities. Considering  these two factors reduces the cost of the humanoid robot drastically, however, it will decrease the controllability of the locomotion and manipulation because of considerable deflection of the links of the robot and the lack of force feedback. The deflection of the link causes a high amount of error for the foot position and orientation and this error leads to the high impact of the foot on the ground and as a consequence falling of the robot. One solution for tackling this problem after impact is using a balance recovery algorithm \cite{shafiee2019online,shafiee2017push, shafiee2017robust} based on estimated CoM\cite{alai2018new}, however, it's reasonable to avoid impact beforehand instead of using a balance recovery after impact.  For solving this problem, 
 we propose a novel and cheap predictive foot sensor based on incremental encoders and a control framework for adapting the configuration of the swing foot for a smooth landing. This foot sensor facilitates dynamically locomotion with a high amount of error in the foot position and orientation as well as enabling the robot to walk on unknown obstacles.

For the manipulation task, controlling the force for grasping different objects with different stiffness is important, and SURENA does not include any Force/Torque sensor in hands. For tackling these challenges, we have designed and implemented a hand based on the current feedback for grasping objects with different stiffness so that the robot is able to drill a hole and write his name on the white board.

\section{Mechanical Design}
\subsection{Design Concepts OF SURENA \rom{4}  Humanoid Robot}
SURENA \rom{4} has been developed with the goal of establishment of key technologies to have a cost-effective humanoid robot for real-world scenarios. The main design concepts of SURENA \rom{4} are: 

\begin{itemize}
\item Lower price
\item Minimizing of the overall mass 
\item Center of mass(CoM) height as high as possible
\item Anthropomorphic design
\end{itemize}

In order to realize the cost-effective feature, we didn't use any force/torque sensor in the robot that are expensive. Instead of that, we used the available facilities in the university that are not the most advanced tools. Therefore, the manufactured parts and connections have deflection causes high amount of error in the foot trajectory in swing mode. To tackle this challenge, we designed a novel and inexpensive predictive foot sensor that enables the robot to walk under this disturbances. To minimize the mass and inertia, we have developed a compact actuator module. The details of these design concepts have been elaborated in the next sections.


\begin{figure}[]
\centering
    \includegraphics[width=\linewidth]{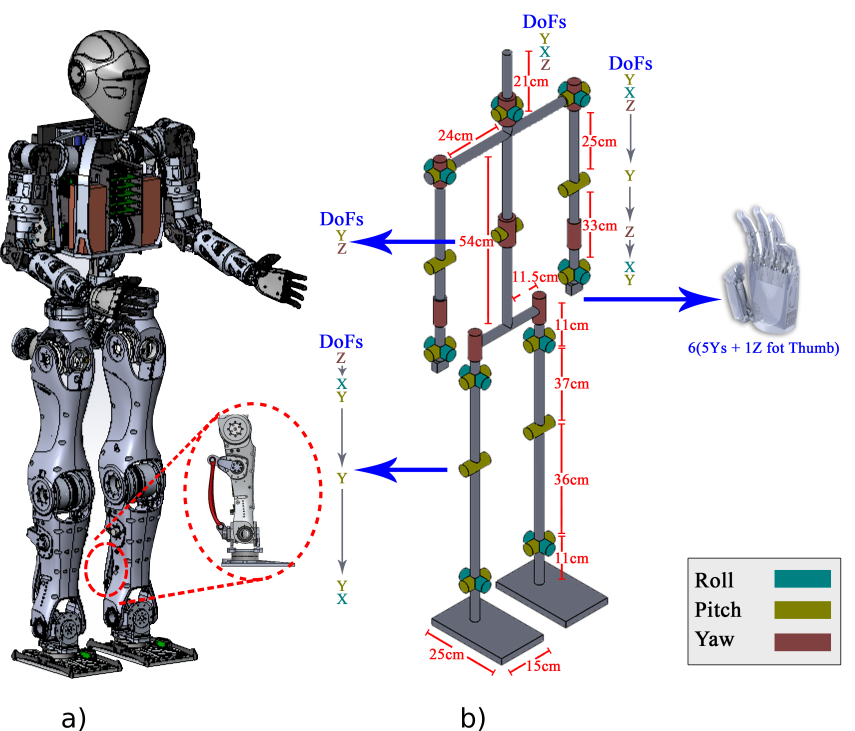}
\caption{The design of the configuration of SURENA \rom{4} a) The structure and components design b) The kinematic structure and degrees of freedom }
      \label{fig:kinematicstructure}
\end{figure}

\begin{table}[h]
\caption{Overview of SURENA\hspace{0.07cm}\rom{4} joint specifications.(Y=Yaw, R=Roll, P=Pitch)}
\label{table_1}
\begin{center}
\begin{tabular}{|c||c||c||c||c||c|}
\hline
\textbf{joint} & \textbf{motor} & \textbf{ratio} & \textbf{${\dot q}_{max}$}&  \textbf{${\tau}_{max}$}&\textbf{range $^\circ$} \\
\hline
hip R & $ RD50\times14$  & 120 & 25 rpm& 60 Nm &-25..30\\
\hline
hip P & $RD50\times14$ & 80 & 38 rpm&40 Nm &$\pm 90$ \\
\hline
hip Y &$ RD50\times14$ & 120 & 25 rpm&60 Nm &$\pm 45$\\
\hline
knee P & RD450 & 50 & 60 rpm&72 Nm& 0..135 \\
\hline
ankle R & $RD50\times08$ & 100 & 50 rpm&27 Nm &$\pm 25$ \\
\hline
ankle P & $RD50\times08$ &100 & 50 rpm&27 Nm& -75..40 \\
\hline
waist P & $RD70\times18$ &100 & 20 rpm& 120 Nm & -10..20 \\
\hline
waist Y &  MX106 & - & 45 rpm&8 Nm& -15..15 \\
\hline
shoulder P & $RD50\times08$ & 120 & 50 rpm&35 Nm& -90..45 \\
\hline
shoulder R & $RD50\times08$   &120 & 50 rpm&35 Nm& -10..90 \\
\hline
shoulder Y & $RD38\times06$   &100 & 100 rpm&10 Nm& -45..45 \\
\hline
elbow P & $RD38\times06$  &100 & 100 rpm&10 Nm& -90..-5 \\
\hline
wrist Y & MX106 & - & 45 rpm& 8 Nm& -60..60 \\
\hline
wrist R& MX106 & - & 45 rpm& 8 Nm& -25..25 \\
\hline
wrist P & MX106 & - & 45 rpm& 8 Nm& -25..25 \\
\hline
neck Y & MX106 & - & 45 rpm& 8 Nm& -60..60 \\
\hline
neck R& MX106 & - & 45 rpm& 8 Nm& -15..15 \\
\hline
neck P & MX106 & - & 45 rpm& 8 Nm& -15..25 \\
\hline
fingers & PQ12 &- & 10 mm/s &50 N& - \\
\hline

\end{tabular}
\end{center}
\end{table}

\begin{figure*}[h]
\centering
\includegraphics[scale=0.8, trim ={0.0cm 0.0cm 0.0cm 0.0cm},clip]{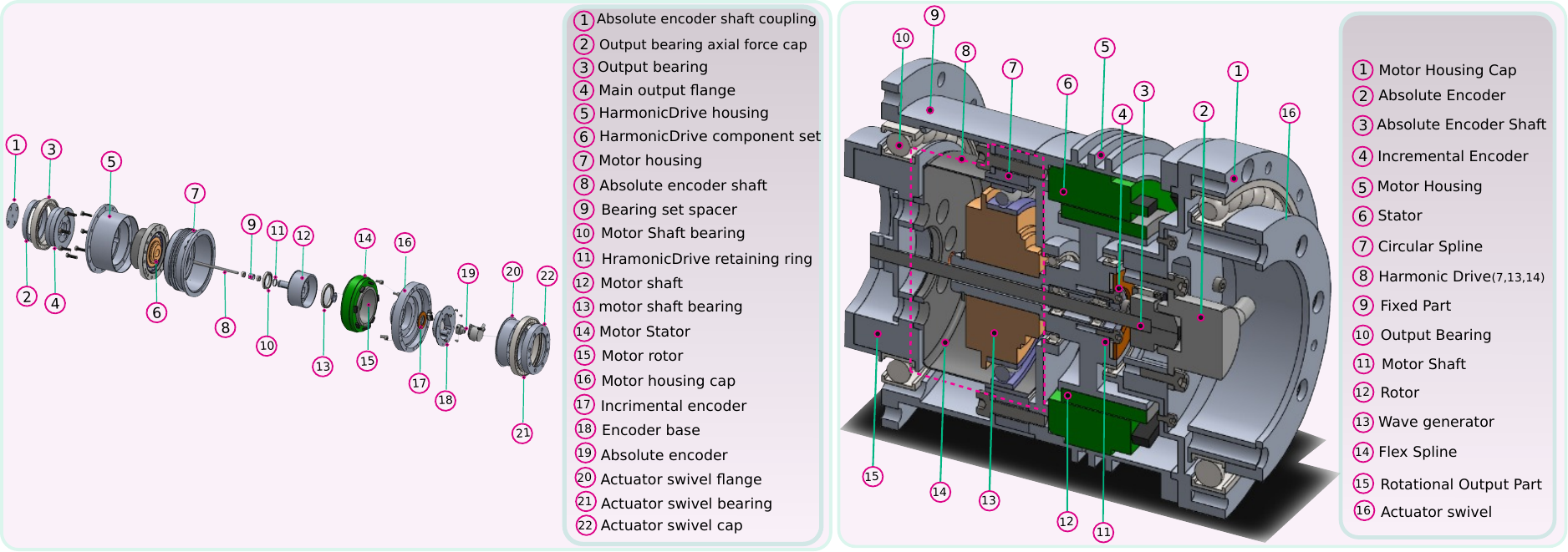}
\caption{Driving unit integrating motor, harmonic-drive gear, incremental and absolute position sensors. (here: Knee joint pitch axis)}
     \label{fig:Actuator}
\end{figure*}

\subsection{Lower-body}
In design process of the legs, the important requirement is having highly integrated module with the anthropomorphic characteristic in addition to low mass and inertia and yet stiff structure. The legs of the SURENA \rom{4} humanoid robot are designed with 6 DoFs at each leg. The mechanical and kinematic structure of the leg has been shown in the Fig. \ref{fig:kinematicstructure}. There are 3 DoFs in the hip, 1 in the knee and 2 in the ankle. The structural parts are made of an magnesium alloy (AZ91) and optimized using FEM-based optimization and iterative process to reduce the weight of the robot and having minimum deflection at the same time. 

We used a four bar mechanism for the ankle pitch axis to bring up the CoM as much as possible  Fig.\hspace{0.05cm}\ref{fig:kinematicstructure}. This mechanism allowed for a relatively slim ankle design and decreased the shank's inertia. The ankle pitch actuator is mounted right below the knee, which allows for a comparably slim and strong ankle design and decreases the leg's inertia. 

We identified the required joint torques and velocities by performing standard motions (e.g. walking under different velocities) in simulations and accordingly selected the actuation parameters (i.e. choice of motors and gears). Table.\hspace{0.05cm}\ref{table_1}  summarizes SURENA IV’s joint specifications.



\subsection{Actuator design}
In order to meet the requirements of legged locomotion, the development of compact and lightweight joint drives is crucial. We use BLDC brush-less motors, because of their high power-to-weight ratio. The motors come as kit motors, which provides a space and weight-saving integration into the joint. The joints employ Harmonic-Drive gears as a gearbox. Their advantages are well known and include no-backlash and high reduction ratios at a low weight. The compact design of Harmonic-Drive component sets allows a space-saving integration. All harmonic-drives are custom lightweight versions with a T-shaped Circular Spline which is, in our experience, the best trade-off between weight and loading capacity. The Wave Generators, modified for low  inertia and weight, are made from aluminum or steel. Angular measurement is done incrementally on motor-side for commutation and again absolutely on the load-side for calibration of the robot. Fig.\hspace{0.05cm}\ref{fig:Actuator} shows the components and design structure of the knee joint pitch axis drive.

\subsection{Upper-body}
In order to perform the required tasks, the following necessary features for upper-body of SURENA \rom{4} are considered: 
\begin{itemize}
\item Lower price
\item Ability to grasp various objects
\item Ability to make human like gestures
\item Light weight
\item Anthropomorphic design
\end{itemize}

\begin{figure}[h]
\centering
\includegraphics[scale=0.55, trim ={0.0cm 0.0cm 0.0cm 0.0cm},clip]{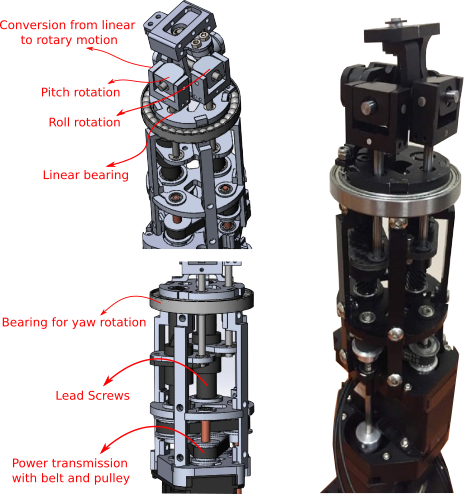}
\caption{The novel wrist mechanism to replicate the spherical joint}
     \label{fig:wrist}
\end{figure}

The upper body of the robot consists of the arm, hand and from waist to head. In total, it consists of 31 DoFs. Each arm has 7 DoFs which make it redundant and each hand includes 6 DoFs, two of which is for thumb and one for each finger. In addition, the waist has 2 DoFs and neck has 3 DoFs as it has been illustrated in Fig. \ref{fig:kinematicstructure}. In the following, we describe the arm and hand design in more details. 
 \begin{itemize}
\item Arm design
\end{itemize}
The arm, along with the forearm, has 7 degrees of freedom. Shoulder roll and shoulder pitch actuator are located on the shoulder joint, shoulder yaw actuator is attached along the arm which exerts minimum load on it, although remained intersected with the other two. Therefore, at the end of the actuator, secondary support is used to tolerate the torque produced during rotational motion of the actuator. Elbow has a rotation about pitch axis which is quite clear. However, the forearm is the challenging part. As shown in Fig.\hspace{0.05cm}\ref{fig:wrist} the challenge is to design a 3-DoF wrist while keeping their axis intersected. The problem has been solved by converting rotary motion into a linear motion and then convert it into rotary motion about a perpendicular axis. Combination of rotation about pitch and roll axes makes the wrist joint act like a spherical joint with minimum clearance. In the whole arm structure, the combination of aluminum alloy and carbon fiber composite is used. 



\begin{figure}[h]
\centering
\includegraphics[scale=0.5, trim ={0.cm 0.0cm 0.0cm 0.0cm},clip]{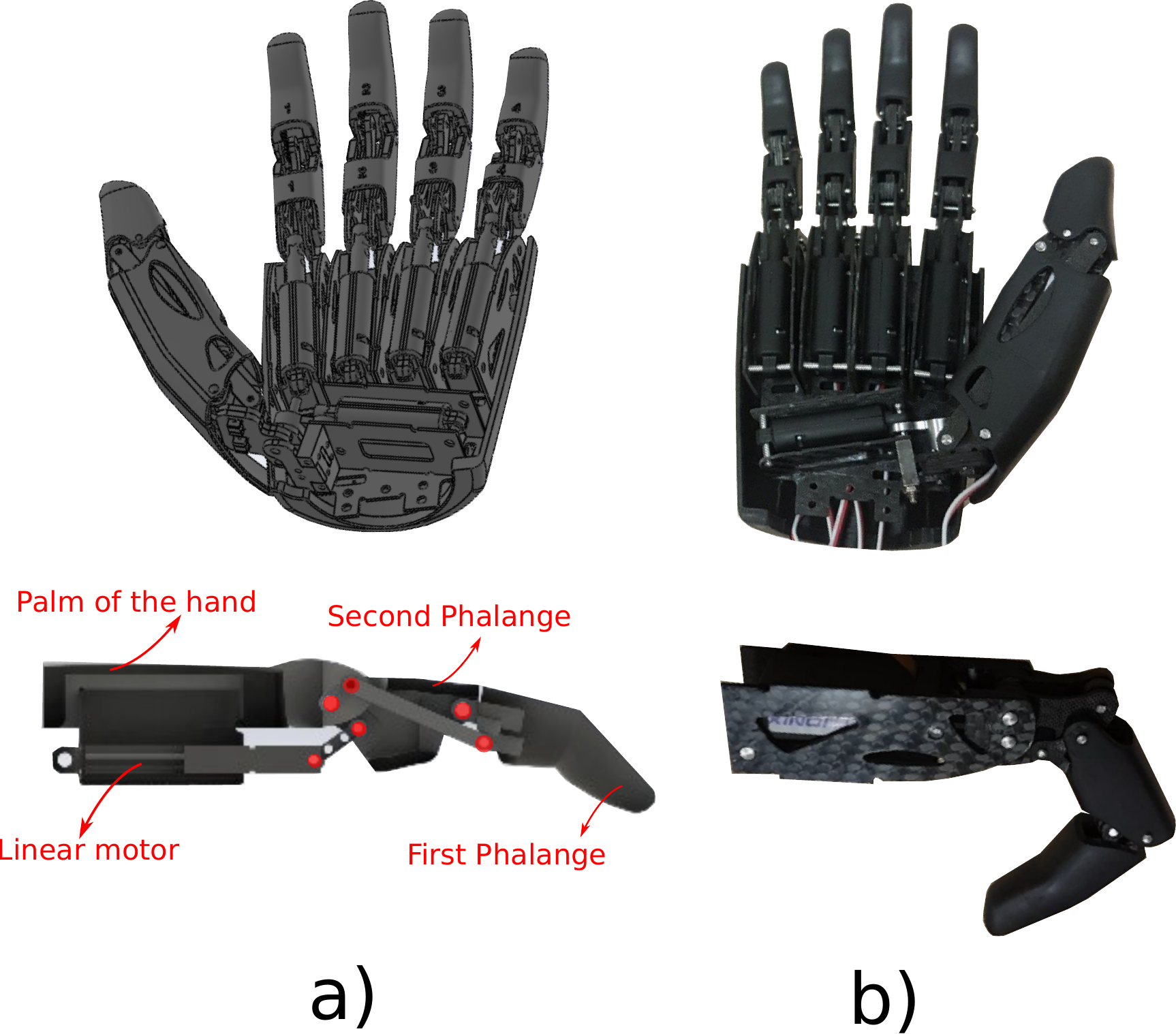}
\caption{The design and fabrication of the hand and fingers of the robot}
     \label{fig:handDesign}
\end{figure}

 \begin{itemize}
\item Hand design
\end{itemize}
Each robot’s hand has been designed to have a similar shape and performance to human. It has six independent degrees of freedom. This design is done in a modular manner so that each
finger is an independent unit which can be removed and repaired. In this case, for the purpose of reducing
the number of communication wires, an intermediate PCB is used at the back of the hand, in which all sensors
and actuators of the hand are connected to it and position control of the fingers which ends up in making different gestures is done by the micro controller mounted on it. In addition, six current sensors are used to manage force control process by a threshold that is set for grasping objects with different stiffness. This process has been implemented by a micro controller which is mounted on this local board.  Fig.\hspace{0.05cm}\ref{fig:handDesign} shows the design of the hand and the 6-bar mechanism of the fingers in SURENA \rom{4} humanoid robot.

\begin{figure}[]
\centering
    
\includegraphics[scale=0.5, trim ={0.2cm 0.0cm 0.0cm 0.0cm},clip]{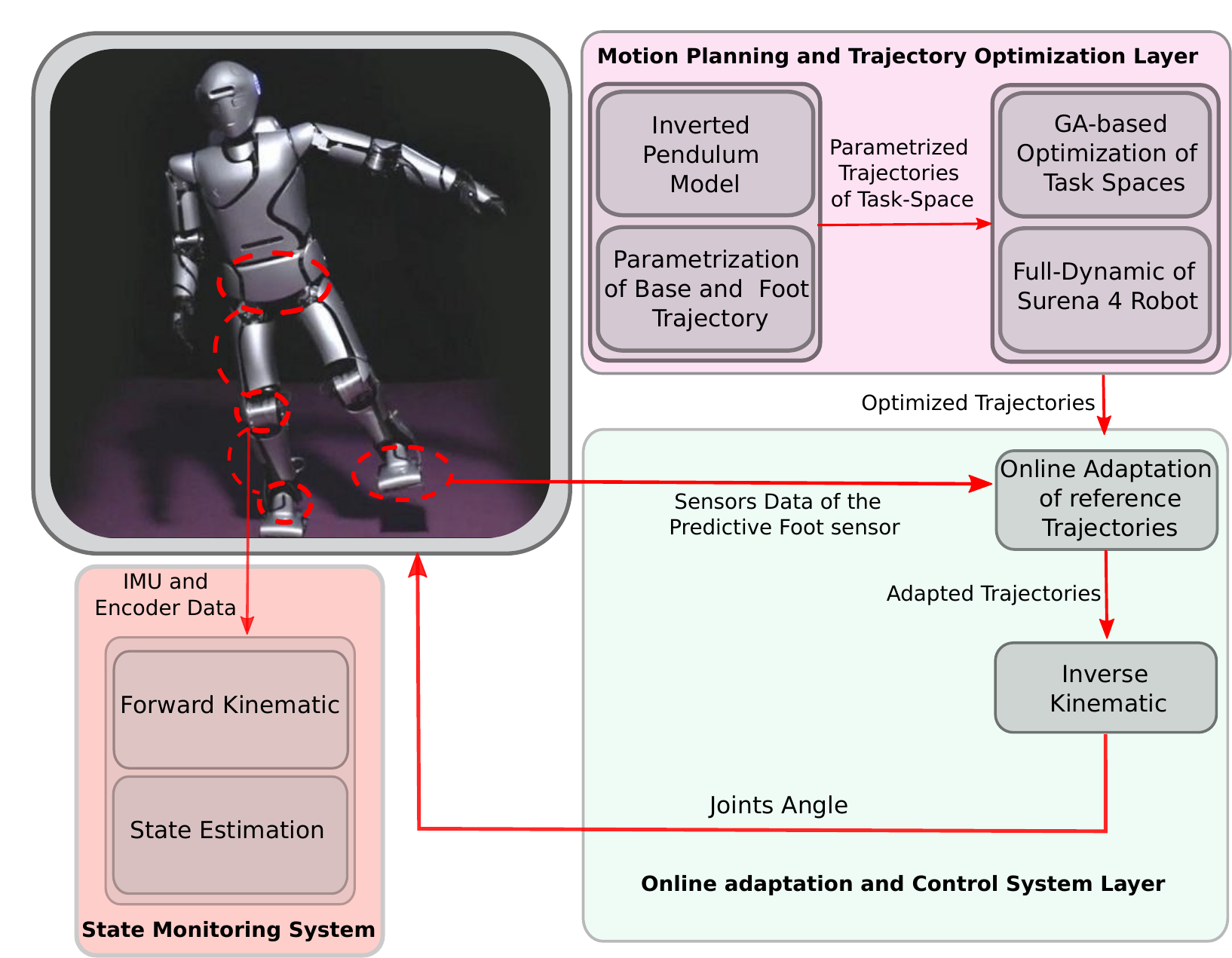}
    
\caption{The motion planning and control architecture}
      \label{fig:control}
\end{figure}

\section{MOTION PLANNING AND CONTROL}
The Fig. \ref{fig:control} shows the proposed control architecture developed for SURENA \rom{4}. It includes different layers that are described in the following sections.

    
    


\subsection{Motion Planning and Trajectory Optimization Layer}
We propose a walking pattern generator for the SURENA \rom{4} humanoid robot that enables the robot to walk dynamically with lower peaks in the torque profile of the ankle motors, which allows us using smaller motors with the purpose of making the robot similar to the human morphology. 

While various motion planning methods are developed based on full-dynamic equations of the robot, the results of most of these planners are smooth trajectories for the CoM but not a smooth motion for the pelvis of the robot, and the motor trajectories calculated based on these methods would include sharp changes that would be challenging for the low-level PID controllers to accurately follow them. This might cause unwanted task-space trajectories or leads to failure. Our method is based on the parameterization of the task-space by polynomials to guarantee smooth motion of the robot, and the coefficients are calculated by optimization of the ZMP derived from the full-dynamic equation. Solving Inverse Kinematics between the desired planned motions for the feet and the pelvis in each iteration gives the desired angles of the joints.

\begin{figure}[h]
\centering
\includegraphics[scale=0.350, trim ={0.0cm 0.0cm 0.0cm 0.0cm},clip]{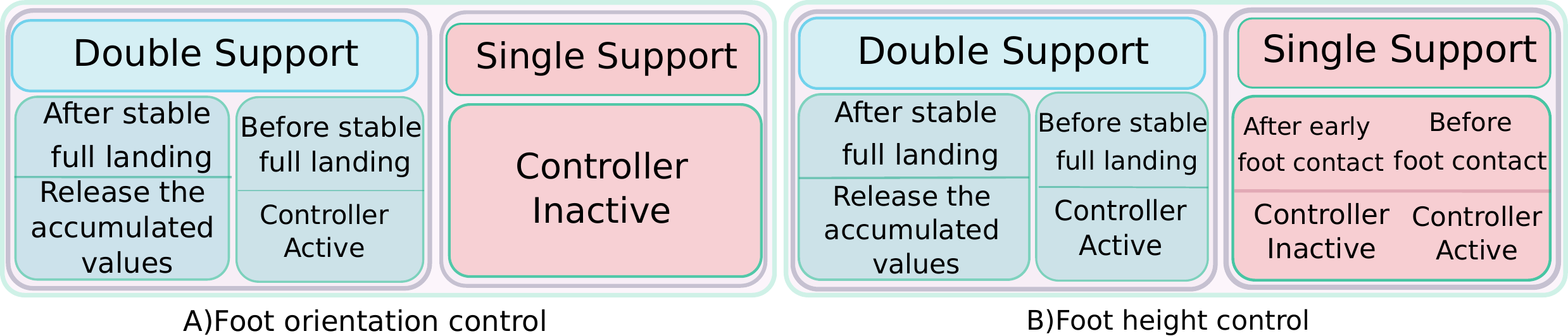}
\caption{The different control and  phases of the height and orientation adaptation of the swing foot based on the predictive foot sensor}
     \label{fig:sensorPhases}
\end{figure}

\begin{figure}[]
\centering
    \includegraphics[scale=0.5, trim ={0.0cm 0.0cm 0.0cm 0.0cm},clip]{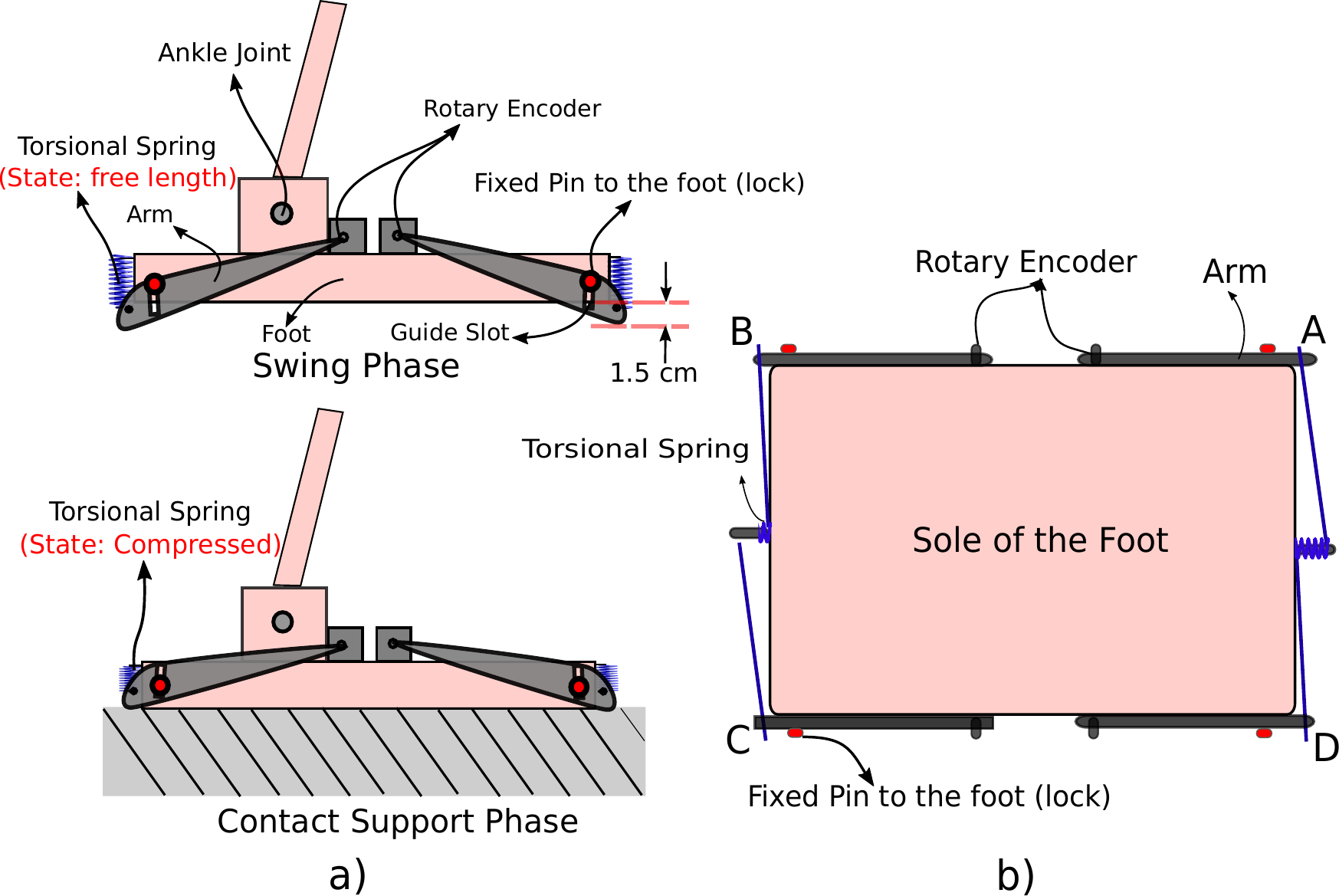}
\caption{ The predictive foot sensor can predict the contact 1.5cm before contact event a)Side view b)Bottom view}
      \label{fig:footSensorSchematic}
\end{figure}


\subsection{Online Adaptation and Control System Layer}
Some major challenges in controlling cost-effective humanoid robots are the presence of deflections in parts and clearances in joints, especially in the SS phase. In Fig. \ref{fig:deflection}, deflections and clearances of the SURENA \rom{4} humanoid robot are shown. These phenomena are common among humanoid robots produced in Universities, due to the low-cost manufacturing process and materials. During the SS phase, the swing leg acts as a cantilever beam, and the early collision of the swinging foot with the ground makes fast walking a challenge even on flat terrain. 

The humanoid robots should also be able to autonomously walk on uneven terrain, especially when the variations are small and hard to be detected by LiDAR sensors or Cameras. A novel controller method alongside new-inexpensive contact sensors in the foot is exploited to effectively overcome the aforementioned challenges. The controller is able to modify both position and orientation of the swinging foot during the last moments of the SS phase (Fig. \ref{fig:footSensorSchematic}), to avoid the severe collision of the foot with the ground during dynamic walking as follows:
\begin{itemize}
\item Four rotary contact sensors are used on the four corners of each foot as described in Fig. \ref{fig:footSensorSchematic}. These sensors are attached to encoders at their end, and the distance of the four corners of the swinging foot to the ground would be reported accurately when they are less than 1.5 cm as it is depicted in Fig. \ref{fig:footSensorSchematic}. Therefore, the average distance of the four corners and the relative pitch and roll angles of the foot to the ground are fed back into the controlling algorithm:

\begin{equation}
d_{avg} = \frac{d_A + d_B + d_C + d_D}{4} 
\label{eq:eq11}
\end{equation}
\begin{equation}
Pitch_{avg,rel}: \phi_{avg} \propto [\frac{d_B + d_C}{2}-\frac{d_A + d_D}{2}]/2
\label{eq:eq12}
\end{equation}
\begin{equation}
Roll_{avg,rel}: \alpha_{avg} \propto [\frac{d_A + d_B}{2}-\frac{d_C + d_D}{2}]/2
\label{eq:eq13}
\end{equation}

\begin{figure}[h]
\centering
\includegraphics[scale=0.5, trim ={0.0cm 0.0cm 0.0cm 0.0cm},clip]{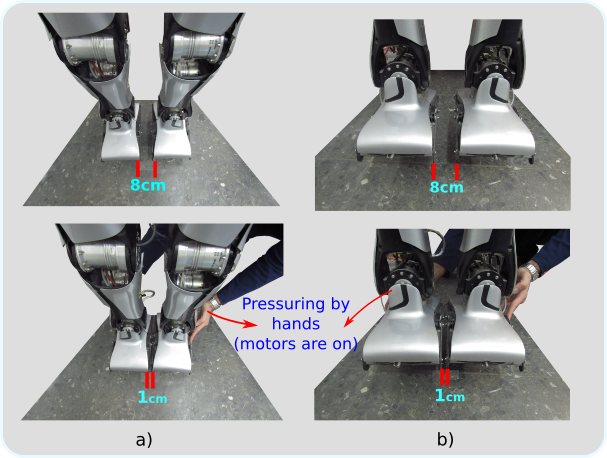}
\caption{The deflection and clearance of the connections leads to about 7 cm error in swing foot motion while the motors are on and tracking the zero position, a)Top view, b)Front view. The accompanying video illustrates this problem.}
     \label{fig:deflection}
\end{figure}

\begin{figure}[]
\centering
    \includegraphics[scale=0.8, trim ={0.0cm 0.0cm 0.0cm 0.0cm},clip]{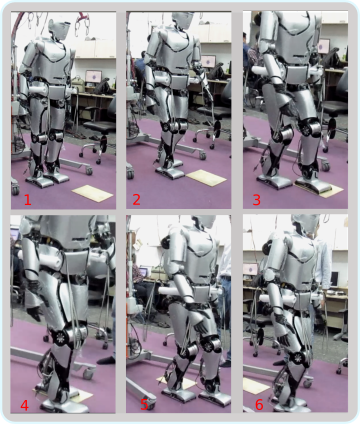}
\caption{Walking on the unknown obstacle with a height of 1cm. The accompanying video demonstrates this capability.}
      \label{fig:walk}
\end{figure}

\item
Due to the deflections in parts and clearances in the joints (as shown in Fig. \ref{fig:deflection}), the relative angle between the swinging foot and the ground would have errors during the SS phase. The calculated average feedback of the relative angles is used in a cumulative controller to make the foot angles follow the desired values by directly manipulating the ankle joints angles (pitch and roll) trajectories. In each iteration (k), the cumulative errors are calculated as follows:
\begin{equation}
\phi_{CE}(k + 1) = \phi_{CE}(k) + [\phi_{Des}(k+1)-\phi_{Avg}(k+1)] 
\label{eq:eq14}
\end{equation}
\begin{equation}
\alpha_{CE}(k + 1) = \alpha_{CE}(k) + [\alpha_{Des}(k+1)-\alpha_{avg}(k+1)]
\label{eq:eq15}
\end{equation}

By simply adding the cumulative error to the desired Pitch and Roll angles in the ankle joint, the input of the low-level controller modified as follows:
\begin{equation}
\alpha_{input}(k+1) = \alpha_{des}(k+1) + \alpha_{CE} (k+1)  \label{eq:eq16}
\end{equation}
\begin{equation}
\phi_{input}(k+1) = \phi_{des}(k+1) + \phi_{CE} (k+1)  \label{eq:eq17}
\end{equation}

\item It is important to note that the effects of deflections and clearances disappear during the DS phase, and the swinging foot would act as the stance foot in the next step. The changes in pitch and roll angles of the ankle joint should be reduced to zero to prevent falling in future steps. This happens during the initial moments of the DS phase when the robot is stable and no longer requires the additional angles provided by the controller. The timeline schematic of the ankle controller is provided in Fig. \ref{fig:sensorPhases}.
\item The early collision of the swinging foot with the ground happens due to the deflections in parts and clearances in the joints. Unlike the angle controller, the position controller manipulates the planned height motion of the swinging foot (z-direction) to reduce the shocking effect of severe collision during dynamic walking. A small threshold (20\%) is also considered to give the controller enough space and time to lay down the foot. When $d_{avg}$ reaches the threshold earlier than expected by the planned trajectories, the controller stops the foot and modifies the planned motion with new polynomials to make the foot slowly lay down on the ground.
\end{itemize}

The experiments show that SURENA \rom{4} is able to autonomously walk on surfaces with bounded variations in elevation (1cm) and slope ($3^{\circ}$) with its highest speed. The snapshots of the robot walking on uneven terrain are shown in Fig. \ref{fig:walk}.


\begin{figure}[h]
\centering
\includegraphics[scale=0.5, trim ={0.0cm 0.0cm 0.0cm 0.0cm},clip]{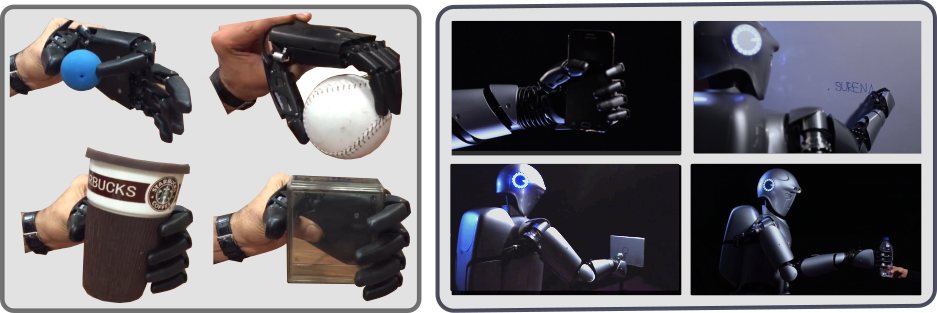}
\caption{Left: Gripping  objects with the different geometries and stiffness. Right: Different tasks by using the combination of whole-body motion generation, gripping and visual perception(The accompanying video shows the performance). }
     \label{fig:manipulation}
\end{figure}

\subsection{Upper-body Motion Generation and Control}

We defined the visual servoing task that robots detects and follows the position of an object (i.e. water bottle) with camera on his head and tracks the position of the bottle by arm to reach the bottle and then proceed with mentioned gripping algorithm to grasp the object. In addition, the robot is able to write his name on the white board thanks to the anthropomorphic design of the hand and arms in addition to the whole-body inverse-kinematic controller.
In Fig.\ref{fig:manipulation}, the examples of manipulation tasks is shown.

In addition, the grasping of the objects with different stiffness and geometries has been shown in Fig. \ref{fig:manipulation}.

\section{DEVELOPMENT OF SOFTWARE AND ELECTRONICS ARCHITECTURE}
\begin{figure}[h]
\centering
\includegraphics[scale=0.5, trim ={0.0cm 0.0cm 0.0cm 0.0cm},clip]{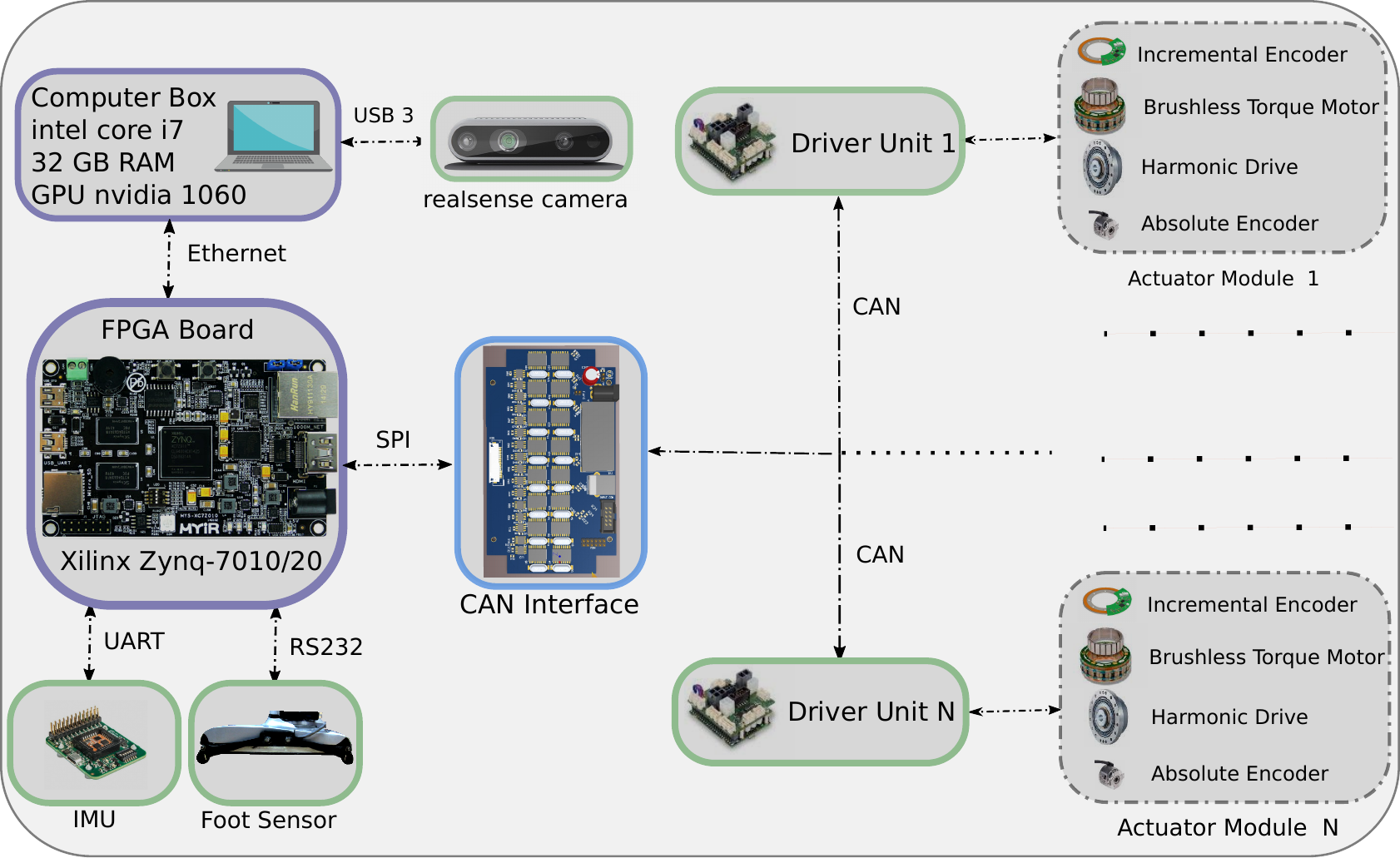}
\caption{The electronic and communications}
      \label{fig:electronic}
\end{figure}
Fig. \ref{fig:electronic} shows the electrical system and the communication protocols that have been used in SURENA\rom{4} humanoid robot. The electronic system consists of two major parts, the main processor and the FPGA board. The main processor generates and sends data to actuators through the FPGA board. Afterward, the FPGA board collects data from sensors and encoders that indicate the position of each actuator and sends them back to the main processor. The total delay of this procedure is 5 milliseconds. Hence, the frequency of executing the control loop is 200 Hz.  

\subsection{Main processor}
The main processor is a computer box that includes a core i7 CPU, 32 GB RAM, and a 1060 NVidia GPU. This computer is responsible for decision making, generating movement data, image and audio processing.

 \subsection{FPGA board}
The FPGA board connects the main processor to actuators and sensors. This board consists of a Xilinx ZYNQ-7020 that include FPGA and two ARM Cortex A9 processors. Using FGPA allows us to communicate with a large number of interfaces simultaneously. The connection between the main processor and the FPGA board is through a Gigabit Ethernet via UDP protocol. The connection between the FPGA board and the motor driver is established via CAN protocol. We used a separate CAN controller to drive each motor by the CAN interface to reach real-time communication.

\subsection{Software architecture}
Communication between different parts of the control application is accomplished by the Robot Operating System (ROS) framework. The SURENA \rom{4} control system consists of four main ROS nodes: simulation node, motor control node, motion planner node, and state estimation node. 

\begin{figure}[h]
\centering
    \includegraphics[scale=0.43, trim ={0.0cm 0.0cm 0.0cm 0.0cm},clip]{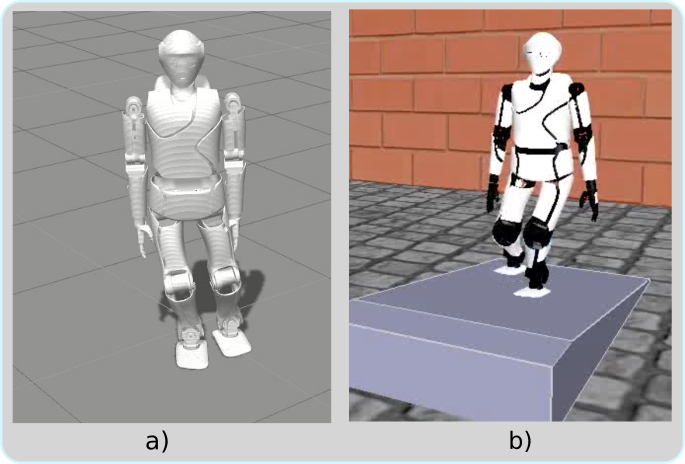}
\caption{Simulation environments:a) Simulation of SURENA walking in Gazebo, b) Simulation of SURENA  slope climbing  in the Choreonoid}
      \label{fig:simulation}
\end{figure}

Several simulators are available under ROS such as  Gazebo which is  widely used, however, Gazebo is relying on the ODE dynamic engine that based on our experience is not the best simulator for the biped robot walking that depend on dynamic contact switching. As it has been illustrated in the Fig. \ref{fig:simulation} the simulation has been done in both Choreonoid and Gazebo. In our experience, Choreonoid that is based on the AistSimulator\cite{nakaoka2012choreonoid} Dynamic Engine  provides a better result and more meaningful simulations with respect to the real experiment.  

\section{Conclusions}
In this paper, we have presented a cost-effective humanoid robot with an anthropomorphic design. The robot components have been manufactured by the university facilities where connections and links have deflections and clearance that causes a high amount of error of about 7 cm of the foot trajectory tracking in task-space in swing mode.This will cause severe impact with ground and falling will happen as a consequence. We presented a novel predictive foot sensor in addition to a full dynamic optimization-based motion planning that enables not only walking with a high amount of foot trajectory error but also locomotion on uneven terrain without any force feedback. This feature facilitates the development of a humanoid robot with non-expensive and simple available manufacturing tools and still robust to a high amount of unknown uncertainty that reduces drastically the price of developing humanoid robots. In addition, we presented an anthropomorphic hand and arm design which enable the robot to do different tasks such as 
grasping objects with different stiffness and geometries, drilling, visual servoing of moving objects, and writing his name on the white-board without any force feedback and only by current feedback of the customized hand actuators.

 \section*{ACKNOWLEDGMENT}
The authors would like to thank the industrial design and artificial intelligence groups respectively for developing  the cover and the perception of the robot.


\bibliography{bibliography}

\begin{thebibliography}{10}
\providecommand{\url}[1]{#1}
\csname url@rmstyle\endcsname
\providecommand{\newblock}{\relax}
\providecommand{\bibinfo}[2]{#2}
\providecommand\BIBentrySTDinterwordspacing{\spaceskip=0pt\relax}
\providecommand\BIBentryALTinterwordstretchfactor{4}
\providecommand\BIBentryALTinterwordspacing{\spaceskip=\fontdimen2\font plus
\BIBentryALTinterwordstretchfactor\fontdimen3\font minus
  \fontdimen4\font\relax}
\providecommand\BIBforeignlanguage[2]{{%
\expandafter\ifx\csname l@#1\endcsname\relax
\typeout{** WARNING: IEEEtran.bst: No hyphenation pattern has been}%
\typeout{** loaded for the language `#1'. Using the pattern for}%
\typeout{** the default language instead.}%
\else
\language=\csname l@#1\endcsname
\fi
#2}}

\bibitem{jung2018development}
T.~Jung, J.~Lim, H.~Bae, K.~K. Lee, H.-M. Joe, and J.-H. Oh, ``Development of
  the humanoid disaster response platform drc-hubo+,'' \emph{IEEE Transactions
  on Robotics}, vol.~34, no.~1, pp. 1--17, 2018.

\bibitem{Kaneko2011}
K.~Kaneko, F.~Kanehiro, M.~Morisawa, K.~Akachi, G.~Miyamori, A.~Hayashi, and
  N.~Kanehira, ``{Humanoid robot HRP-4 - Humanoid robotics platform with
  lightweight and slim body},'' in \emph{IEEE Int. Conf. Intell. Robot. Syst.},
  2011.

\bibitem{kaneko2019humanoid}
K.~Kaneko, H.~Kaminaga, T.~Sakaguchi, S.~Kajita, M.~Morisawa, I.~Kumagai, and
  F.~Kanehiro, ``Humanoid robot hrp-5p: An electrically actuated humanoid robot
  with high-power and wide-range joints,'' \emph{IEEE Robotics and Automation
  Letters}, vol.~4, no.~2, pp. 1431--1438, 2019.

\bibitem{kojima2015development}
K.~Kojima, T.~Karasawa, T.~Kozuki, E.~Kuroiwa, S.~Yukizaki, S.~Iwaishi,
  T.~Ishikawa, R.~Koyama, S.~Noda, F.~Sugai, \emph{et~al.}, ``Development of
  life-sized high-power humanoid robot jaxon for real-world use,'' in
  \emph{2015 IEEE-RAS 15th International Conference on Humanoid Robots
  (Humanoids)}.\hskip 1em plus 0.5em minus 0.4em\relax IEEE, 2015, pp.
  838--843.

\bibitem{Tsagarakis2007}
N.~G. Tsagarakis, G.~Metta, G.~Sandini, D.~Vernon, R.~Beira, F.~Becchi,
  L.~Righetti, J.~Santos-Victor, A.~J. Ijspeert, M.~C. Carrozza, and D.~G.
  Caldwell, ``{ICub: The design and realization of an open humanoid platform
  for cognitive and neuroscience research},'' \emph{Adv. Robot.}, 2007.

\bibitem{Tsagarakis2017}
N.~G. Tsagarakis, D.~G. Caldwell, F.~Negrello, W.~Choi, L.~Baccelliere, V.~G.
  Loc, J.~Noorden, L.~Muratore, A.~Margan, A.~Cardellino, L.~Natale, E.~{Mingo
  Hoffman}, H.~Dallali, N.~Kashiri, J.~Malzahn, J.~Lee, P.~Kryczka,
  D.~Kanoulas, M.~Garabini, M.~Catalano, M.~Ferrati, V.~Varricchio,
  L.~Pallottino, C.~Pavan, A.~Bicchi, A.~Settimi, A.~Rocchi, and A.~Ajoudani,
  ``{WALK-MAN: A High-Performance Humanoid Platform for Realistic
  Environments},'' \emph{J. F. Robot.}, 2017.

\bibitem{radford2015valkyrie}
N.~A. Radford, P.~Strawser, K.~Hambuchen, J.~S. Mehling, W.~K. Verdeyen, A.~S.
  Donnan, J.~Holley, J.~Sanchez, V.~Nguyen, L.~Bridgwater, \emph{et~al.},
  ``Valkyrie: Nasa's first bipedal humanoid robot,'' \emph{Journal of Field
  Robotics}, vol.~32, no.~3, pp. 397--419, 2015.

\bibitem{digit}
A.~Aggarwal, ``Digit humanoid robot,''
  \url{https://www.agilityrobotics.com/meet-digit}, May 2020.

\bibitem{englsberger2014overview}
J.~Englsberger, A.~Werner, C.~Ott, B.~Henze, M.~A. Roa, G.~Garofalo, R.~Burger,
  A.~Beyer, O.~Eiberger, K.~Schmid, \emph{et~al.}, ``Overview of the
  torque-controlled humanoid robot toro,'' in \emph{2014 IEEE-RAS International
  Conference on Humanoid Robots}.\hskip 1em plus 0.5em minus 0.4em\relax IEEE,
  2014, pp. 916--923.

\bibitem{stasse2017talos}
O.~Stasse, T.~Flayols, R.~Budhiraja, K.~Giraud-Esclasse, J.~Carpentier,
  J.~Mirabel, A.~Del~Prete, P.~Sou{\`e}res, N.~Mansard, F.~Lamiraux,
  \emph{et~al.}, ``Talos: A new humanoid research platform targeted for
  industrial applications,'' in \emph{2017 IEEE-RAS 17th International
  Conference on Humanoid Robotics (Humanoids)}.\hskip 1em plus 0.5em minus
  0.4em\relax IEEE, 2017, pp. 689--695.

\bibitem{kajita2014introduction}
S.~Kajita, H.~Hirukawa, K.~Harada, and K.~Yokoi, \emph{Introduction to humanoid
  robotics}.\hskip 1em plus 0.5em minus 0.4em\relax Springer, 2014, vol. 101.

\bibitem{jeong2020design}
H.~Jeong, K.~Lee, W.~Kim, I.~Lee, and J.-H. Oh, ``Design and control of the
  rapid legged platform gazelle,'' \emph{Mechatronics}, vol.~66, p. 102319,
  2020.

\bibitem{buschmann2009humanoid}
T.~Buschmann, S.~Lohmeier, and H.~Ulbrich, ``Humanoid robot lola: Design and
  walking control,'' \emph{Journal of physiology-Paris}, vol. 103, no. 3-5, pp.
  141--148, 2009.

\bibitem{kamioka2018simultaneous}
T.~Kamioka, H.~Kaneko, T.~Takenaka, and T.~Yoshiike, ``Simultaneous
  optimization of zmp and footsteps based on the analytical solution of
  divergent component of motion,'' in \emph{2018 IEEE International Conference
  on Robotics and Automation (ICRA)}.\hskip 1em plus 0.5em minus 0.4em\relax
  IEEE, 2018, pp. 1763--1770.

\bibitem{kheddar2019humanoid}
A.~Kheddar, S.~Caron, P.~Gergondet, A.~Comport, A.~Tanguy, C.~Ott, B.~Henze,
  G.~Mesesan, J.~Englsberger, M.~A. Roa, \emph{et~al.}, ``Humanoid robots in
  aircraft manufacturing: The airbus use cases,'' \emph{IEEE Robotics \&
  Automation Magazine}, vol.~26, no.~4, pp. 30--45, 2019.

\bibitem{kopacek2011cost}
P.~Kopacek, ``Cost oriented humanoid robots,'' \emph{IFAC Proceedings Volumes},
  vol.~44, no.~1, pp. 12\,680--12\,685, 2011.

\bibitem{shafiee2019online}
M.~Shafiee, G.~Romualdi, S.~Dafarra, F.~J.~A. Chavez, and D.~Pucci, ``Online
  dcm trajectory generation for push recovery of torque-controlled humanoid
  robots,'' in \emph{2019 IEEE-RAS 19th International Conference on Humanoid
  Robots (Humanoids)}.\hskip 1em plus 0.5em minus 0.4em\relax IEEE, 2019, pp.
  671--678.

\bibitem{shafiee2017push}
M.~Shafiee-Ashtiani, A.~Yousefi-Koma, R.~Mirjalili, H.~Maleki, and M.~Karimi,
  ``Push recovery of a position-controlled humanoid robot based on capture
  point feedback control,'' in \emph{2017 5th RSI International Conference on
  Robotics and Mechatronics (ICRoM)}.\hskip 1em plus 0.5em minus 0.4em\relax
  IEEE, 2017, pp. 126--131.

\bibitem{shafiee2017robust}
M.~Shafiee-Ashtiani, A.~Yousefi-Koma, and M.~Shariat-Panahi, ``Robust bipedal
  locomotion control based on model predictive control and divergent component
  of motion,'' in \emph{2017 IEEE International Conference on Robotics and
  Automation (ICRA)}.\hskip 1em plus 0.5em minus 0.4em\relax IEEE, 2017, pp.
  3505--3510.

\bibitem{alai2018new}
H.~Alai, F.~A. Shirazi, and A.~Yousefi-Koma, ``New approach to center of mass
  estimation for humanoid robots based on sensor measurements and general
  lipm,'' in \emph{2018 6th RSI International Conference on Robotics and
  Mechatronics (IcRoM)}.\hskip 1em plus 0.5em minus 0.4em\relax IEEE, 2018, pp.
  388--393.

\bibitem{nakaoka2012choreonoid}
S.~Nakaoka, ``Choreonoid: Extensible virtual robot environment built on an
  integrated gui framework,'' in \emph{2012 IEEE/SICE International Symposium
  on System Integration (SII)}.\hskip 1em plus 0.5em minus 0.4em\relax IEEE,
  2012, pp. 79--85.

\end{thebibliography}
\bibliographystyle{IEEEtran}

\end{document}